\documentclass[final]{cvpr}

\usepackage{times}
\usepackage{epsfig}
\usepackage{graphicx}
\usepackage{amsmath}
\usepackage{amssymb}
\usepackage{makecell}
\usepackage{bbding}
\usepackage{tablefootnote}

\usepackage[pagebackref=true,breaklinks=true,colorlinks,bookmarks=false]{hyperref}

\def\cvprPaperID{****} 
\def\confYear{CVPR 2022}

\begin{document}

\title{Winning the CVPR'2022 AQTC Challenge:\\ A Two-stage Function-centric Approach}


\author{Shiwei Wu \and
Weidong He \and
Tong Xu \and
Hao Wang \and
Enhong Chen \\
Anhui Province Key Laboratory of Big Data Analysis and Application, \\ State Key Laboratory of Cognitive Intelligence, \\University of Science and Technology of China \\
{\{dwustc, hwd\}@mail.ustc.edu.cn}
{\{tongxu, wanghao3, cheneh\}@ustc.edu.com}
}

\maketitle

\begin{abstract}
   Affordance-centric Question-driven Task Completion for Egocentric Assistant~(AQTC) is a novel task which helps AI assistant learn from instructional videos and scripts and guide the user step-by-step. In this paper, we deal with the AQTC via a two-stage Function-centric approach, which consists of Question2Function Module to ground the question with the related function and Function2Answer Module to predict the action based on the historical steps. We evaluated several possible solutions in each module and obtained significant gains compared to the given baselines. Our code is available at \url{https://github.com/starsholic/LOVEU-CVPR22-AQTC}.
\end{abstract}

\section{Introduction}
Intelligent assistants are increasingly becoming a part of users' daily lives. Along this line, Affordance-centric Question-driven Task Completion ~(AQTC)~\cite{wong2022assistq}, which aims to guide the user to deal with unfamiliar events step-by-step with the knowledge learned from instructional videos and scripts, is newly introduced. Different from existing works such as Visual Question Answering~(VQA)~\cite{antol2015vqa} or Visual Dialog~\cite{das2017visual}, the question in AQTC is about specific task and the answer is multi-modal and multi-step, which makes it more challenging. 

To solve this problem, we propose a novel two-stage Function-centric approach, which consists of a Question2Function Module and a Function2Answer Module. Our main motivation is that only part of the instructional video is helpful to answer the question and taking the entire video into account could introduce unnecessary noise. Along this line, we first define several schemas to segment the scripts into the textual function-paras. Then we design a text similarity based method to select specific video clips as well as paras that are closely related to users' question. After obtaining relative context information, we formulate the multi-step QA as a classification task and leverage a neural network to retrieve correct answer for each step.

With our model and several training tricks, we achieved substantial performance boost compared to the given baselines.

\section{Related Work}
\subsection{Measurement of Text Similarity}
Generally speaking, in order to calculate text similarity, it is important to represents the text as numerical features that can be calculated directly, which could be categorized into two groups: string-based method and corpus-based method.

String-based methods aim to measure similarity between two text strings based on string sequences or character composition, including character-based methods~\cite{irving1992two,winkler1990string} and phrase-based methods~\cite{dice1945measures,jaccard1912distribution}. Different from string-based methods, corpus-based methods leverage the textual feature or co-occurrence probability to calculate the text similarity at the corpus level, which are usually achieved in three ways: bag-of-words model like Term Frequency–Inverse Document Frequency~(TF-IDF)~\cite{robertson1994some}, distributed representation methods like Word2Vec~\cite{mikolov2013efficient} and BERT~\cite{devlin2018bert}, and matrix factorization methods like Latent Semantic Analysis~(LSA)~\cite{deerwester1990indexing} and Latent Dirichlet Allocation~(LDA)~\cite{blei2003latent}.

\subsection{Visual Question Answering}
VQA is a task to answer questions based on image~\cite{vqa} or video~\cite{tvqa}, which could be roughly divided into attention based methods~\cite{butd, san} and bilinear pooling based approaches~\cite{mcbp, ban, mfbp}. \cite{butd} developed different attention modules to adaptively attend on the relevant image regions based on the question representation. \cite{mcbp} proposed to employ the compact bilinear pooling methods to combine the visual and linguistic features. However, these tasks mainly focus on the third-person perspective, while the AQTC task concentrates on the egocentric scenes.

\section{Proposed Method}
\begin{figure*}
\centering
\includegraphics[width=1.03\textwidth]{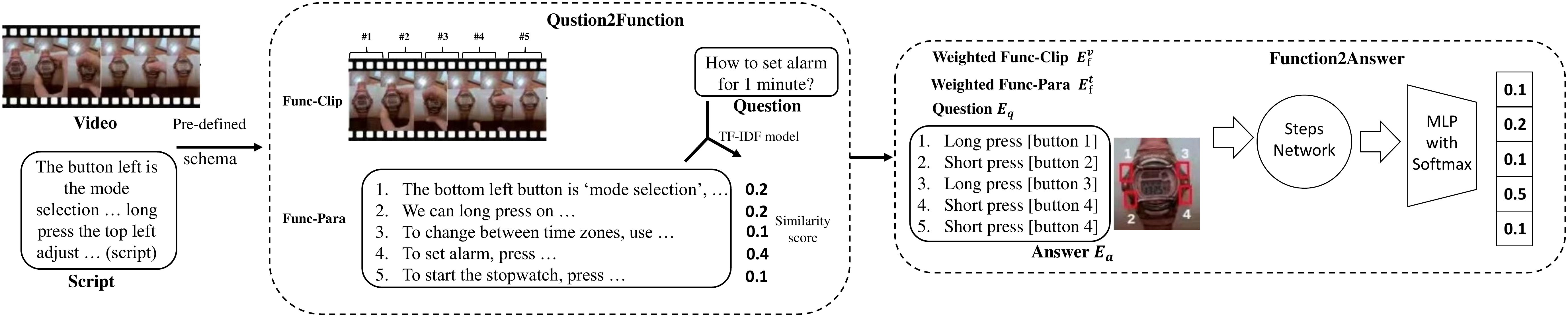}
\caption{The Proposed Function-centric approach}
\label{fig:framework}
\end{figure*}

In this section, we first present the problem statement of our proposed two-stage Function-centric approach towards solving the AQTC task. Then, we introduce the technical details of each module in our framework step-by-step as shown in Fig.\ref{fig:framework}.

\subsection{Problem Statement}
Given an instructional video $V$, the video's corresponding textual script $S$, the user's question $Q$, the set of candidate answers $A^i_j$ where $A^i_j$ denotes the $j$-th potential answer in the $i$-th step, we target at select one correct answer in each step. Specifically, to grounding question $Q$ in the instructional video $V$ and script $S$, we first segment $V$ and $S$ into function set $\{f_1, f_2, \dots, f_n\}$ and then matching $Q$ with related functions. Note that each function $f$ consists of function-clip $f^v$ and function-para $f^t$. Afterwards, taking the weighted function set $\{f_1, f_2, \dots, f_n\}$, the question $Q$ and the candidate answer $A^i_j$ as input, we formulate the multi-step QA as a classification task in a supervised way.

\subsection{Question2Function Module}
We now turn to explain the technical details for the Question2Function Module. Since the instructional videos are used to guide user or AI assistant in a step-by-step manner, we first segment both script and video into individual functions, instead of sentences or frames, to insure the completeness of each step. Meanwhile, it is critical to ground the specific question with the related function as the correct answer often co-occurs with the corresponding function.    

Specifically, we first segment the script $S$ into the textual function-paras $f^t$ according to the pre-defined schema~(see details in our project's repo), and then divide the corresponding video $V$ into the visual function-clips $f^v$ via the aligned script timestamp. In this way, the instructional video and script are divided into the functions set$\{f_1, f_2, \dots, f_n\}$, and each function $f$ not only contains textual description $f^t$ but also includes visual guidance $f^v$. 

We further match the specific question with the function set based on text similarity. Due to the small volume of the dataset and the corresponding functions are not highly semantic similar with the given question, we calculate the similarity score between $Q$ and $\{f_1, f_2, \dots, f_n\}$ via TF-IDF model~\cite{tfidf} instead of deep learning based methods. The ablation result in Sec.\ref{ablation} indicates that the traditional statistical based TF-IDF method behaves much better than the deep learning based approach.


\subsection{Function2Answer Module}
After we determined the related function thanks to the former module, we now turn to formulate the following multi-step QA as a classification task. Since the candidate answers are given as textual action descriptions and visual button images, we need to predict the correct action in each step well as the corresponding button according to the historical steps. 

\textbf{Input Features.} For the text embedding, we encode the function-para, question and candidate answers into $E_f^t$, $E_q^t$ and $E_a^t$ via XL-Net~\cite{xlnet}, which performs much better than the Bert~\cite{bert} backbone as shown in Tab.\ref{table1}, since the XL-Net is good at processing long context. For the visual part, we encode the frames of function-clip and the button image in candidate answers into $E_f^v$ and $E_a^v$ via vision transformer~(ViT)~\cite{vit} following ~\cite{assistq}.

\textbf{Steps Network and Prediction Head.} Same as baseline~\cite{assistq}, we use GRU~\cite{gru} to leverage the historical steps and predict the final score for each answer via a two-layer MLP followed by softmax activation.

\textbf{Loss Function.} By taking the embeddings of question $E_q^t$, candidate answer $E_a=\{E_a^t, E_a^v\}$ and the weighted function set $E_f=\{E_f^t, E_f^v\}$ given by the former module as input, we formulate the following multi-step QA as a classification task as follows.

\begin{equation}
    L = \sum\limits_{i=1}^{N} -y_i \log \hat{y_i},  \\
\end{equation}
where $\hat{y_i}$ = Pred\_head(Steps\_network($E_f$, $E_q^t$, $E_a$)) and $y_i$ represents the ground truth.

\subsection{Other Attempts}
Considering the changing views within the same instructional video and the occlusion of buttons, it is really challenging to link the button's image to its function. Therefore, we tried to use finger detection~\cite{fingerdet} in the video as additional information in the Function2Answer module~(see details in our project's repo). However, since the strong assumption and the cumulative error introduced by the detection module, the performance of this purely-inference method is poor as shown in Tab.\ref{table2}.

\section{Experiments}
\subsection{Dataset and Parameters setting}
Following ~\cite{assistq}, we trained our model on the training set containing 80 instructional videos and evaluated the model performance on the testing set which contains 20 instructional videos. We use the same evaluation metrics in~\cite{assistq}, i.e., Recall@k, Mean rank~(MR) and Mean reciprocal rank~(MRR).

For fair comparison, we train our model as well as ablation experiments under the same parameters setting~(Adam optimizer with the learning rate $1 \times 10^{-4} $, maximum 100 training epochs and 16 batchsize) and report the best evaluation metrics epoch on the testing set.

\subsection{Ablation Study} \label{ablation}
In this section, we compare the performance of possible solutions in each module separately. 

\begin{table}[!h]
\centering
\setlength{\tabcolsep}{1.5mm}{
\begin{tabular}{c|c|c|c|c|c}
    \hline
     & \makecell{Grounding\\ Approach} & R@1 & R@3 & MR & MRR \\ 
     \hline
    \makecell{Baseline with \\ raw settings} & cross-att. & 30.2 & 62.3 & 3.2 & 3.2 \\
    \hline
    \makecell{Sentence-\\centric} & \makecell{cross-att. \\TF-IDF} & \makecell{39.3 \\ 44.4} & \makecell{70.2 \\ 75.0} & \makecell{2.8 \\ 2.7} & \makecell{3.5 \\ 3.8}\\  
    \hline
    \makecell{Function-\\centric} & \makecell{cross-att. \\TF-IDF} & \makecell{38.5 \\ \textbf{45.2}} & \makecell{74.2 \\ \textbf{75.4}} & \makecell{2.9 \\ \textbf{2.6}} & \makecell{3.5 \\ \textbf{3.9}}\\
    \hline
\end{tabular}}
\vspace{3pt}
\caption{Ablation studies in Question2Function module}
\label{table1}
\end{table}

\vspace{-3pt}

\begin{figure}[!h]
\centering
\includegraphics[width=0.5\textwidth]{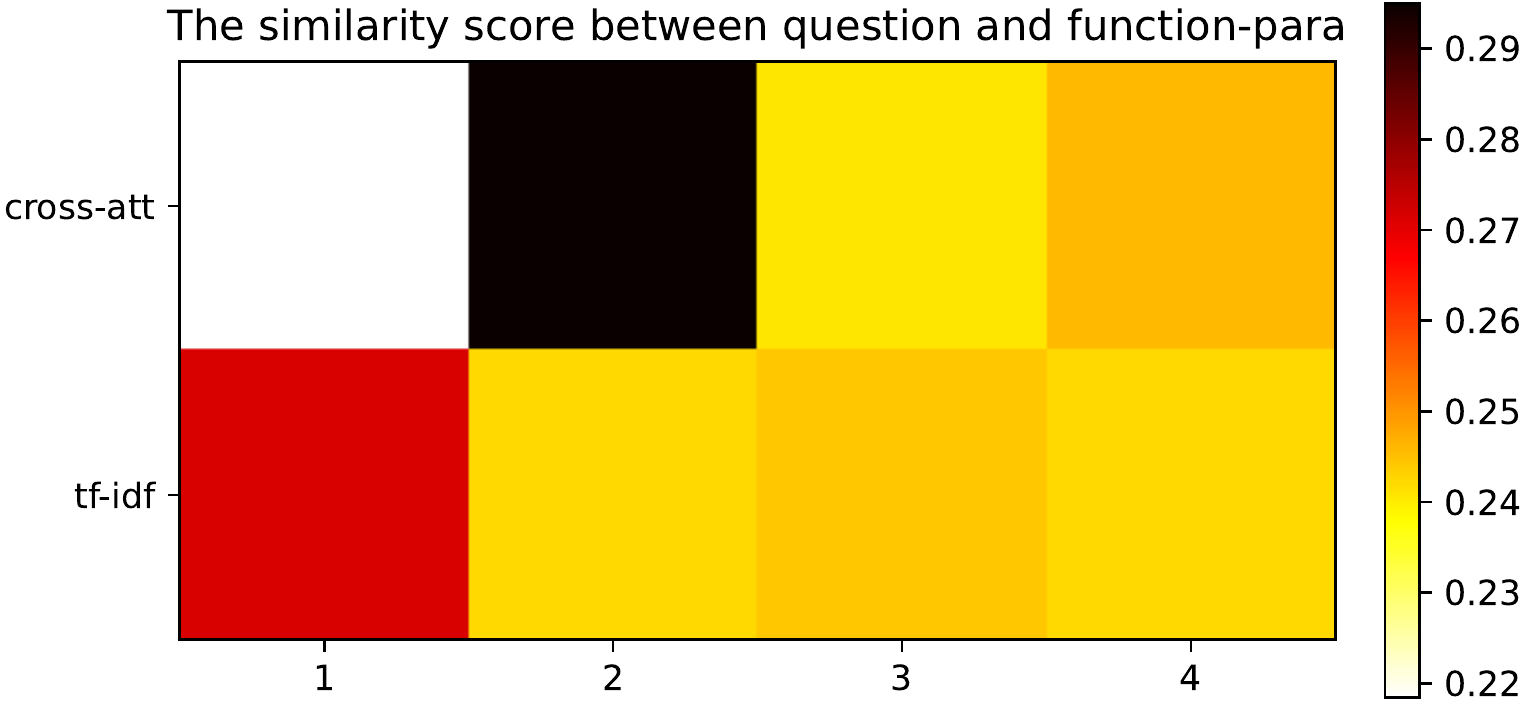}
\caption{The similarity score between question and function-para. \\ \textbf{Question}: How to defrost 2kg of fish? \\ \textbf{Function-paras}: $\#1$ How to defrost 1kg of fish? Press turbo defrost button. Turn the time knob clockwise to 1 kg. Press the start button. \\$\#2$ How to microwave eggplant at medium high power for 30 seconds? Press micropower button twice. Turn time knob clockwise to 30 seconds. start button. \\ $\#3$ How to set microwave to 1 minute timer? Turn time knob clockwise to 1 minute. Press the start button. How to stop microwave in the middle of use? Press the sensor reheat button. Press the start button. \\$\#4$ To stop, press the stop button. To resume, press the start button again. How to start, stop, start and stop microwave? Press the sensor reheat button.Press the start button. To stop, press the stop button. To resume, press the start button again. To stop, press the stop button.}
\label{fig:attention}
\end{figure}

Specifically, in the Question2Function module, we evaluate the different segment methods~(sentence-centric v.s. function-centric) as well as the grounding approaches~(cross attention v.s. TF-IDF) as shown in Tab.\ref{table1}. Compared with baseline using raw settings, we observe that the adjustment of the text encoder(BERT $\rightarrow$ XL-Net) and optimizer~(SGD $\rightarrow$ Adam) has a significant impact on performance~(30.2$R@1$ $\rightarrow$ 39.3$R@1$). Meanwhile, the function-centric segment approach performs better than the sentence-centric on most of the metrics, which demonstrates the significance of the step completeness. For grounding approach, the TF-IDF model performs much better than the cross-attention mechanism~(39.3$R@1$ $\rightarrow$ 44.4$R@1$ on sentence-centric and 38.5$R@1$ $\rightarrow$ 45.2$R@1$ on function-centric). We also show an case~(see Fig.\ref{fig:attention}) between the two different grounding approaches, which reveals that TF-IDF model can help us find the related functions effectively.

For Function2Answer module, we evaluate the impact of visual and textual part of functions and answers via extensive ablation studies. As we can see, the visual guidance of the function helps the model choose the correct answer significantly~(41.0$R@1$ $\rightarrow$ 45.2$R@1$). However, the inclusion of the visual part of answer brings little gains on metrics~(43.3$R@1$ $\rightarrow$ 45.2$R@1$), which reveals that the link of button image to its function should be further investigated.

\vspace{-3pt}
\begin{table}[!h]
\vspace{-0.4em}
    \centering
    \begin{tabular}{c|c|c|c|c|c}
        \hline
        \makecell{Functions \\ V T} & \makecell{Answers \\ V T} & R@1 & R@3 & MR & MRR \\
        \hline
        \makecell{\Checkmark \Checkmark} & \makecell{\XSolidBrush \Checkmark} & 43.3 & \textbf{76.6} & 2.6 & 3.8 \\ 
        \makecell{\XSolidBrush \Checkmark} & \makecell{\Checkmark \Checkmark} & 41.0 & 71.8 & 2.8 & 3.7 \\ 
        \makecell{\Checkmark \Checkmark} & \makecell{\Checkmark \Checkmark} & \textbf{45.2} & 75.4 & \textbf{2.6} & \textbf{3.9} \\ 
        \hline
        \multicolumn{2}{c|}{Finger detection~\cite{fingerdet}} & 23.8 & 66.8 & 3.1 & 3.0 \\
        \hline
        \multicolumn{2}{c|}{\makecell{Submitted version \\ on CodaLab~\tablefootnote{https://codalab.lisn.upsaclay.fr/competitions/4642\#results}}} & 41.0 & 72.0 & 2.8 & 3.7 \\ 
        \hline
    \end{tabular}
    \vspace{3pt}
    \caption{Ablation studies in Function2Answer module}
    \label{table2}
\end{table}


\section{Conclusion}
In this paper,  we introduced a two-stage method to solve the novel AQTC task. Our model achieved significant performance boost compared to the given baseline. For future work, we will attempt other solutions to model this interesting task.

\section{Acknowledgement}
This work was supported by the grants from the National Natural Science Foundation of China (No.62072423).

{\small
\bibliographystyle{ieee_fullname}
\bibliography{egbib}

\begin{thebibliography}{10}\itemsep=-1pt

\bibitem{fingerdet}
Mohammad~Mahmudul Alam, Mohammad~Tariqul Islam, and SM~Mahbubur Rahman.
\newblock Unified learning approach for egocentric hand gesture recognition and
  fingertip detection.
\newblock {\em Pattern Recognition}, 121:108200, 2022.

\bibitem{butd}
Peter Anderson, Xiaodong He, Chris Buehler, Damien Teney, Mark Johnson, Stephen
  Gould, and Lei Zhang.
\newblock Bottom-up and top-down attention for image captioning and visual
  question answering.
\newblock In {\em 2018 {IEEE} Conference on Computer Vision and Pattern
  Recognition, {CVPR} 2018, Salt Lake City, UT, USA, June 18-22, 2018}, pages
  6077--6086. Computer Vision Foundation / {IEEE} Computer Society, 2018.

\bibitem{antol2015vqa}
Stanislaw Antol, Aishwarya Agrawal, Jiasen Lu, Margaret Mitchell, Dhruv Batra,
  C~Lawrence Zitnick, and Devi Parikh.
\newblock Vqa: Visual question answering.
\newblock In {\em Proceedings of the IEEE international conference on computer
  vision}, pages 2425--2433, 2015.

\bibitem{vqa}
Stanislaw Antol, Aishwarya Agrawal, Jiasen Lu, Margaret Mitchell, Dhruv Batra,
  C.~Lawrence Zitnick, and Devi Parikh.
\newblock {VQA:} visual question answering.
\newblock In {\em 2015 {IEEE} International Conference on Computer Vision,
  {ICCV} 2015, Santiago, Chile, December 7-13, 2015}, pages 2425--2433. {IEEE}
  Computer Society, 2015.

\bibitem{blei2003latent}
David~M Blei, Andrew~Y Ng, and Michael~I Jordan.
\newblock Latent dirichlet allocation.
\newblock {\em Journal of machine Learning research}, 3(Jan):993--1022, 2003.

\bibitem{gru}
Junyoung Chung, {\c{C}}aglar G{\"{u}}l{\c{c}}ehre, KyungHyun Cho, and Yoshua
  Bengio.
\newblock Empirical evaluation of gated recurrent neural networks on sequence
  modeling.
\newblock {\em CoRR}, abs/1412.3555, 2014.

\bibitem{das2017visual}
Abhishek Das, Satwik Kottur, Khushi Gupta, Avi Singh, Deshraj Yadav,
  Jos{\'e}~MF Moura, Devi Parikh, and Dhruv Batra.
\newblock Visual dialog.
\newblock In {\em Proceedings of the IEEE conference on computer vision and
  pattern recognition}, pages 326--335, 2017.

\bibitem{deerwester1990indexing}
Scott Deerwester, Susan~T Dumais, George~W Furnas, Thomas~K Landauer, and
  Richard Harshman.
\newblock Indexing by latent semantic analysis.
\newblock {\em Journal of the American society for information science},
  41(6):391--407, 1990.

\bibitem{bert}
Jacob Devlin, Ming{-}Wei Chang, Kenton Lee, and Kristina Toutanova.
\newblock {BERT:} pre-training of deep bidirectional transformers for language
  understanding.
\newblock In Jill Burstein, Christy Doran, and Thamar Solorio, editors, {\em
  Proceedings of the 2019 Conference of the North American Chapter of the
  Association for Computational Linguistics: Human Language Technologies,
  {NAACL-HLT} 2019, Minneapolis, MN, USA, June 2-7, 2019, Volume 1 (Long and
  Short Papers)}, pages 4171--4186. Association for Computational Linguistics,
  2019.

\bibitem{devlin2018bert}
Jacob Devlin, Ming-Wei Chang, Kenton Lee, and Kristina Toutanova.
\newblock Bert: Pre-training of deep bidirectional transformers for language
  understanding.
\newblock {\em arXiv preprint arXiv:1810.04805}, 2018.

\bibitem{dice1945measures}
Lee~R Dice.
\newblock Measures of the amount of ecologic association between species.
\newblock {\em Ecology}, 26(3):297--302, 1945.

\bibitem{vit}
Alexey Dosovitskiy, Lucas Beyer, Alexander Kolesnikov, Dirk Weissenborn,
  Xiaohua Zhai, Thomas Unterthiner, Mostafa Dehghani, Matthias Minderer, Georg
  Heigold, Sylvain Gelly, Jakob Uszkoreit, and Neil Houlsby.
\newblock An image is worth 16x16 words: Transformers for image recognition at
  scale.
\newblock In {\em 9th International Conference on Learning Representations,
  {ICLR} 2021, Virtual Event, Austria, May 3-7, 2021}. OpenReview.net, 2021.

\bibitem{mcbp}
Akira Fukui, Dong~Huk Park, Daylen Yang, Anna Rohrbach, Trevor Darrell, and
  Marcus Rohrbach.
\newblock Multimodal compact bilinear pooling for visual question answering and
  visual grounding.
\newblock In Jian Su, Xavier Carreras, and Kevin Duh, editors, {\em Proceedings
  of the 2016 Conference on Empirical Methods in Natural Language Processing,
  {EMNLP} 2016, Austin, Texas, USA, November 1-4, 2016}, pages 457--468. The
  Association for Computational Linguistics, 2016.

\bibitem{irving1992two}
Robert~W Irving and Campbell~B Fraser.
\newblock Two algorithms for the longest common subsequence of three (or more)
  strings.
\newblock In {\em Annual Symposium on Combinatorial Pattern Matching}, pages
  214--229. Springer, 1992.

\bibitem{jaccard1912distribution}
Paul Jaccard.
\newblock The distribution of the flora in the alpine zone. 1.
\newblock {\em New phytologist}, 11(2):37--50, 1912.

\bibitem{ban}
Jin{-}Hwa Kim, Jaehyun Jun, and Byoung{-}Tak Zhang.
\newblock Bilinear attention networks.
\newblock In Samy Bengio, Hanna~M. Wallach, Hugo Larochelle, Kristen Grauman,
  Nicol{\`{o}} Cesa{-}Bianchi, and Roman Garnett, editors, {\em Advances in
  Neural Information Processing Systems 31: Annual Conference on Neural
  Information Processing Systems 2018, NeurIPS 2018, December 3-8, 2018,
  Montr{\'{e}}al, Canada}, pages 1571--1581, 2018.

\bibitem{tvqa}
Jie Lei, Licheng Yu, Mohit Bansal, and Tamara~L. Berg.
\newblock {TVQA:} localized, compositional video question answering.
\newblock In Ellen Riloff, David Chiang, Julia Hockenmaier, and Jun'ichi
  Tsujii, editors, {\em Proceedings of the 2018 Conference on Empirical Methods
  in Natural Language Processing, Brussels, Belgium, October 31 - November 4,
  2018}, pages 1369--1379. Association for Computational Linguistics, 2018.

\bibitem{mikolov2013efficient}
Tomas Mikolov, Kai Chen, Greg Corrado, and Jeffrey Dean.
\newblock Efficient estimation of word representations in vector space.
\newblock {\em arXiv preprint arXiv:1301.3781}, 2013.

\bibitem{tfidf}
Juan Ramos et~al.
\newblock Using tf-idf to determine word relevance in document queries.
\newblock In {\em Proceedings of the first instructional conference on machine
  learning}, volume 242, pages 29--48. Citeseer, 2003.

\bibitem{robertson1994some}
Stephen~E Robertson and Steve Walker.
\newblock Some simple effective approximations to the 2-poisson model for
  probabilistic weighted retrieval.
\newblock In {\em SIGIR’94}, pages 232--241. Springer, 1994.

\bibitem{winkler1990string}
William~E Winkler.
\newblock String comparator metrics and enhanced decision rules in the
  fellegi-sunter model of record linkage.
\newblock 1990.

\bibitem{wong2022assistq}
Benita Wong, Joya Chen, You Wu, Stan~Weixian Lei, Dongxing Mao, Difei Gao, and
  Mike~Zheng Shou.
\newblock Assistq: Affordance-centric question-driven task completion for
  egocentric assistant.
\newblock {\em arXiv preprint arXiv:2203.04203}, 2022.

\bibitem{assistq}
Benita Wong, Joya Chen, You Wu, Stan~Weixian Lei, Dongxing Mao, Difei Gao, and
  Mike~Zheng Shou.
\newblock Assistq: Affordance-centric question-driven task completion for
  egocentric assistant.
\newblock {\em CoRR}, abs/2203.04203, 2022.

\bibitem{xlnet}
Zhilin Yang, Zihang Dai, Yiming Yang, Jaime~G. Carbonell, Ruslan Salakhutdinov,
  and Quoc~V. Le.
\newblock Xlnet: Generalized autoregressive pretraining for language
  understanding.
\newblock In Hanna~M. Wallach, Hugo Larochelle, Alina Beygelzimer, Florence
  d'Alch{\'{e}}{-}Buc, Emily~B. Fox, and Roman Garnett, editors, {\em Advances
  in Neural Information Processing Systems 32: Annual Conference on Neural
  Information Processing Systems 2019, NeurIPS 2019, December 8-14, 2019,
  Vancouver, BC, Canada}, pages 5754--5764, 2019.

\bibitem{san}
Zichao Yang, Xiaodong He, Jianfeng Gao, Li Deng, and Alexander~J. Smola.
\newblock Stacked attention networks for image question answering.
\newblock In {\em 2016 {IEEE} Conference on Computer Vision and Pattern
  Recognition, {CVPR} 2016, Las Vegas, NV, USA, June 27-30, 2016}, pages
  21--29. {IEEE} Computer Society, 2016.

\bibitem{mfbp}
Zhou Yu, Jun Yu, Jianping Fan, and Dacheng Tao.
\newblock Multi-modal factorized bilinear pooling with co-attention learning
  for visual question answering.
\newblock In {\em {IEEE} International Conference on Computer Vision, {ICCV}
  2017, Venice, Italy, October 22-29, 2017}, pages 1839--1848. {IEEE} Computer
  Society, 2017.

\end{thebibliography}
}

\end{document}